\documentclass{IEEEtran}
\IEEEoverridecommandlockouts
\usepackage[hidelinks]{hyperref}
\usepackage{cite}
\usepackage{amsmath,amssymb,amsfonts}
\usepackage{algorithmic}
\usepackage{booktabs}
\usepackage{multirow}
\usepackage{graphicx}
\usepackage{textcomp}
\usepackage{xcolor}
\usepackage{siunitx}
\usepackage{lipsum}
\usepackage[para]{threeparttable}
\usepackage{url}
\usepackage[normalem]{ulem}

\usepackage{subcaption}
\usepackage{dblfloatfix}

\usepackage{mathtools}
\DeclarePairedDelimiter{\ceil}{\lceil}{\rceil}

\usepackage[utf8]{inputenc}
\usepackage[T1]{fontenc}

\usepackage{caption}
\usepackage{tikz}
\usetikzlibrary{shapes,arrows}

\newcommand{\figref}[1]{Fig.~\ref{#1}}
\newcommand{\tblref}[1]{Table~\ref{#1}}
\newcommand{\secref}[1]{Section~\ref{#1}}
\newcommand{\x}{$\times$}
\graphicspath{{./figs/}{./}{./figs/bio/}}

\newcommand{\novel}[1]{{#1}}
\newcommand{\revOne}[1]{{#1}}

\begin{document}
\title{\novel{EBPC: Extended Bit-Plane Compression for Deep Neural Network Inference and Training Accelerators}}
\author{
	Lukas Cavigelli, \IEEEmembership{Member, IEEE}, 
	Georg Rutishauser, \IEEEmembership{Student Member, IEEE}, 
	Luca Benini, \IEEEmembership{Fellow, IEEE}
\thanks{Manuscript received January XX, XXXX; revised August YY, YYYY. The authors would like to thank armasuisse Science \& Technology for funding this research. This project was supported in part by the EU's H2020 programme under grant no. 732631 (OPRECOMP).}
\thanks{L. Cavigelli, G. Rutishauser and L. Benini are with the Integrated Systems Laboratory, ETH Zürich, 8092 Zürich, Switzerland (e-mail: cavigelli@iis.ee.ethz.ch).}
\thanks{L. Benini is also with the Department of Electrical, Electronic and Information Engineering, University of Bologna, 40136 Bologna, Italy.} 
}
\maketitle
\begin{abstract}
\novel{
In the wake of the success of convolutional neural networks in image classification, object recognition, speech recognition, etc., the demand for deploying these compute-intensive ML models on embedded and mobile systems with tight power and energy constraints at low cost, as well as for boosting throughput in data centers, is growing rapidly. This has sparked a surge of research into specialized hardware accelerators. Their performance is typically limited by I/O bandwidth, power consumption is dominated by I/O transfers to off-chip memory, and on-chip memories occupy a large part of the silicon area. 

We introduce and evaluate a novel, hardware-friendly, and lossless compression scheme for the feature maps present within convolutional neural networks. \revOne{We present hardware architectures and synthesis results for the compressor and decompressor in 65\,nm. With a throughput of one 8-bit word/cycle at 600\,MHz, they fit into 2.8\,kGE and 3.0\,kGE of silicon area, respectively---together the size of less than seven 8-bit multiply-add units at the same throughput.} 

We show that an average compression ratio of 5.1\x{} for AlexNet, 4\x{} for VGG-16, 2.4\x{} for ResNet-34 and 2.2\x{} for MobileNetV2 can be achieved---a gain of 45--70\% over existing methods. Our approach also works effectively for various number formats, has a low frame-to-frame variance on the compression ratio, and achieves compression factors for gradient map compression during training that are even better than for inference. 
}

\end{abstract}


\section{Introduction}
\novel{
\IEEEPARstart{C}{omputer} vision has evolved into a key component for automating data analysis over a wide range of field applications: medical diagnostics~\cite{Litjens2017}, industrial quality assurance~\cite{Bian2016}, video surveillance~\cite{Cavigelli2016a}, advanced driver assistance systems~\cite{Wu2016a} and many more. A large number of these applications have only emerged recently due to the tremendous accuracy improvements---even beyond human capabilities~\cite{HePReLU2015}---associated with the advent of deep learning and in particular convolution neural networks (CNNs, ConvNets). 

Even though CNN-based solutions often require considerable computing resources, many of these applications have to run in real-time and on embedded and mobile systems. As a result, purpose-built platforms, application-specific hardware accelerators, and optimized algorithms have been engineered to reduce the number of arithmetic operations and their precision requirements~\cite{Cavigelli2016,Cavigelli2015a,Albericio2016,Parashar2017,Zhang2016c,Aimar2017,Chen2016a,Han2016a,Cavigelli2018,Cavigelli2015}. 

Examining these hardware platforms, the amount of energy required to load and store intermediate results/feature maps (and gradients maps during training) in the off-chip memory is not only significant but typically dominating the energy consumed during computation and on-chip data buffering. This energy bottleneck is even more remarkable when considering networks that are engineered to reduce computing energy by quantizing weights to one or two bits or power-of-two values, dispensing with the need for high-precision multiplications and significantly reducing weight storage requirements~\cite{Andri2018,Andri2016,Courbariaux2015a,Zhou2017a,Andri2018a}. 
}

Many compression methods for CNNs have been proposed over the last few years. However, many of them are focusing exclusively on 
\begin{enumerate}
 \item compressing the parameters/weights, which make up only \revOne{part} of the energy-intensive off-chip communication\revOne{---especially when intermediate results during the computation of a layer are also stored off-chip}~\cite{Chen2015a,Agustsson2017,Han2016,Stock2019}, 
 \item exploiting the sparsity of intermediate results, which is not always present (e.g., in partial results of a convolution layer or otherwise before the activation function is applied) and is not optimal in the sense that the non-uniform value distribution is not capitalized~\cite{Rhu2018,Gudovskiy2018,Wang2017a}, 
 \item very complex methods requiring large dictionaries, or otherwise not suitable for a small, energy-efficient hardware implementation---often targeting efficient distribution and storage of trained models to mobile devices or the transmission of intermediate feature maps from/to mobile devices over a costly communication link~\cite{Han2016}.
\end{enumerate}

\novel{
In contrast to these, the focus on this paper is on reducing the energy consumption of hardware accelerators for CNN inference and training by cutting down on the dominant power contributor---I/O transfers. These data transfers to and from off-chip memory consist of the network parameters (read-only) and the feature maps (read/write). 
\begin{table*}
    \centering
    \caption{\novel{Comparison of Feature Map Sizes and Parameter Count of Modern DNNs}}
    \label{tbl:featureMapsVsParams}
\novel{
\begin{threeparttable}
\begin{tabular}{c|l@{\hskip 1mm}c@{\hskip 2mm}c@{\hskip 1.5mm}c|crr@{\hskip 2mm}r|rrr}
\toprule
 & \multirow{2}{*}{Network} & & Accuracy & \multirow{2}{*}{Dataset} & \multirow{2}{*}{Resolution\tnote{1}} & \multirow{2}{*}{\#MACs} & \multirow{2}{*}{\#params} & \multirow{2}{*}{\#FM values\tnote{2}} & \multicolumn{3}{c}{I/O-ratio FM/parameter\tnote{3}} \\
 & & & [\% top-1/top-5] & & & & & & TWN-infer. & FP-infer. & training   \\
\midrule
    \parbox[t]{2.5mm}{\multirow{6}{*}{\rotatebox[origin=c]{90}{Recognition}}}
    & ResNet-50        & \cite{He2015}      & 77.2 / 93.3  & ILSVRC12 & 224\x 224 & 4.1 G      & 25.6 M   & 11.1 M    &  9.1\x & 0.9\x  &  27.8\x \\
    & DenseNet-121     & \cite{Iandola2014} & 76.4 / 93.3  & ILSVRC12 & 224\x 224 & 2.9 G      &  8.0 M   &  6.9 M    & 17.2\x & 1.7\x  &  55.3\x \\
    & SqueezeNet       & \cite{Iandola2016} & 57.5 / 80.3  & ILSVRC12 & 224\x 224 & 355.9 M    &  1.2 M   &  2.6 M    & 42.4\x & 4.2\x  & 134.1\x \\
    & ShuffleNet2      & \cite{Ma2018}      & 69.4 / ----- & ILSVRC12 & 224\x 224 & 150.6 M    &  2.3 M   &  2.0 M    & 17.2\x & 1.7\x  &  54.8\x \\
    & MobileNet2       & \cite{Sandler2018} & 72.0 / ----- & ILSVRC12 & 224\x 224 & 320.2 M    &  3.5 M   &  6.7 M    & 38.4\x & 3.8\x  & 121.9\x \\
    & MnasNet          & \cite{Tan2018}     & 75.6 / 92.7  & ILSVRC12 & 224\x 224 & 330.2 M    &  4.4 M   &  5.5 M    & 25.2\x & 2.5\x  &  79.7\x \\
    \midrule
    \parbox[t]{2mm}{\multirow{5}{*}{\rotatebox[origin=c]{90}{Detection}}}
    & YOLOv3           & \cite{Redmon2018}  & 57.9\% AP${}_{50}$ & COCO-det.   & 480\x 640 &   9.5 G & 61.6 M  &  68.4 M  & 22.2\x & 2.2\x   &  71.1\x \\
    & YOLOv3-tiny      & \cite{Redmon2018}  & 33.1\% AP${}_{50}$ & COCO-det.   & 480\x 640 & 800.0 M &  8.7 M  &  10.7 M  & 24.2\x & 2.4\x   &  78.4\x \\
    & OpenPose         & \cite{Cao2017}     & 65.3\% mAP         & COCO-keyp.  & 480\x 640 &  50.4 G & 52.3 M  & 132.5 M  & 51.5\x & 5.1\x   & 162.1\x \\
    & MultiPoseNet-50  & \cite{Kocabas2018} & 64.3\% mAP         & COCO-keyp.  & 480\x 640 &  13.3 G & 36.7 M  &  96.0 M  & 52.5\x & 5.2\x   & 167.4\x \\
    & MultiPoseNet-101 & \cite{Kocabas2018} & 62.3\% mAP         & COCO-keyp.  & 480\x 640 &  16.8 G & 55.6 M  & 119.9 M  & 43.4\x & 4.3\x   & 138.0\x \\
\bottomrule
\end{tabular}
\begin{tablenotes}
\item[1] This resolution is used to determine the number of multiply-accumulate operations and feature map values. For the detection CNNs, this differs from the one used during training. 
\item[2] We count the number of feature maps values wherever they are activated (e.g., by a ReLU layer). 
\item[3] Feature maps values are assumed to be 16\,bit and counted twice since they are written and read (most HW accelerator required multiple reads, though). Modes --- \emph{TWN-inference}: batch size 1, ternary weights; \emph{full-precision inference}: batch size 1, 16\,bit weights; \emph{training}: batch size 32, 16\,bit weights. 
\end{tablenotes}
\end{threeparttable}
}

\end{table*}

The latter is the more substantial contributor to the overall transfers as highlighted in \tblref{tbl:featureMapsVsParams}. \revOne{Previous work has further shown that the parameters can even be quantized to ternary representations (1.58\,bit/value) with minimal to no accuracy loss \cite{Zhou2017a, anonymous2020random}, although current commercially available solutions generally target 8-bit quantized weights and activations. As can be seen in our comparison, there seems to be a clear trend  towards applications working with high-resolution data such as object detection are using more I/O for the feature maps than the parameters. The same can be seen with model size and compute effort-optimized networks, and conversely very deep and parameter-rich networks for image classification benefit less. Altogether, this clearly highlights the requirement for feature map compression for energy-constrained scenarios. There, the feature maps outweigh the parameters by 20--50\x{} (16\,bit to 1.58\,bit and 2--5\x{} more values) and the energy share spent on feature map I/O and buffering is growing even more dominant with the simpler arithmetic operations~\cite{Andri2016, Andri2018a, Andri2018, Andri2019} and further work on model compression. }

\paragraph*{Contributions}
\novel{
We are extending our work in \cite{Cavigelli2018a} and make the following main contributions:
\begin{enumerate}
    \item A comparison of state-of-the-art DNNs regarding bandwidth and/or memory size requirements for parameters and feature maps, showing the relevance of feature map compression; 
    \item An in-depth analysis of the feature and gradient maps' properties indicating compressibility; 
    \item The proposal of a novel, state-of-the-art, and hardware-friendly compression scheme; 
    \item A thorough evaluation of its capabilities for inference as well as training (compressing gradient maps); 
    \item A hardware architecture for the compressor and decompressor and a detailed analysis of its implementation in 65\,nm CMOS technology. 
\end{enumerate}
}
\revOne{
This paper is organized as follows: In \secref{sec:related}, we explore previous work on feature map compression in neural networks and the exploitation of feature map sparsity, placing a special focus on hardware suitability. Then, in \secref{sec:algo}, we introduce the \emph{extended bit-plane compression} algorithm and its properties. \secref{sec:hw} describes a hardware implementation of the algorithm and discusses the implementation results. In \secref{sec:results}, detailed results for the application of the algorithm to various networks are presented and discussed, drawing comparisons to other compression methods. \secref{sec:conclusion} concludes the paper, briefly summarizing our results.
}}

\section{Related Work}
\label{sec:related}
\revOne{
We will now introduce some sparsity-based methods for feature map compression which are very hardware friendly and widely used to feed the feature map data in and out of hardware accelerators which use sparsity to reduce to number of compute operations. We then expand the scope of our review to methods targeting model compression and compressed representation learning before taking a more detailed look at a system-level method used to compress feature maps for transfer between GPU and CPU memory. 
}
\subsection{\revOne{Sparsity-based Feature Map Compression}}
There are several \revOne{publications in literature} describing hardware accelerators which exploit feature map sparsity to reduce computation: Cnvlutin \cite{Albericio2016}, SCNN \cite{Parashar2017}, Cambricon-X \cite{Zhang2016c}, NullHop \cite{Aimar2017}, Eyeriss \cite{Chen2016a}, EIE \cite{Han2016a}. Their focus is on power gating or skipping some of the operations and memory accesses. \novel{This entails defining a scheme to feed the data into the system. They all use one of four methods: }
\revOne{
\begin{enumerate}
 \item Zero-RLE (used in SCNN \cite{Parashar2017}): A simple run-length encoding for the zero values, i.e., a single prefix bit followed by the number of zero-values or the non-zero value. 
 \item Zero-free neuron array format (ZFNAf) (used in Cnvlutin \cite{Albericio2016}): Similarly to the widely-used compressed sparse row (CSR) format, non-zero elements are encoded with an offset and their value. 
 \item Compressed column storage (CCS) format (e.g., used in EIE \cite{Han2016a}): Similar to ZFNAf, but the offsets are stored in relative form, thus requiring fewer bits to store them. Few bits are sufficient, and in case they are all exhausted, a zero-value can be encoded as if it was non-zero. 
 \item Zero-value compression (ZVC) (first used in this context in NullHop \cite{Aimar2017}, then in cDMA \cite{Rhu2018}): Saves a fixed-length mask indicating whether a value was zero or non-zero and a variable-length list of the non-zero values. 
\end{enumerate}
}
\begin{figure*}[b]
	\includegraphics[width=\linewidth]{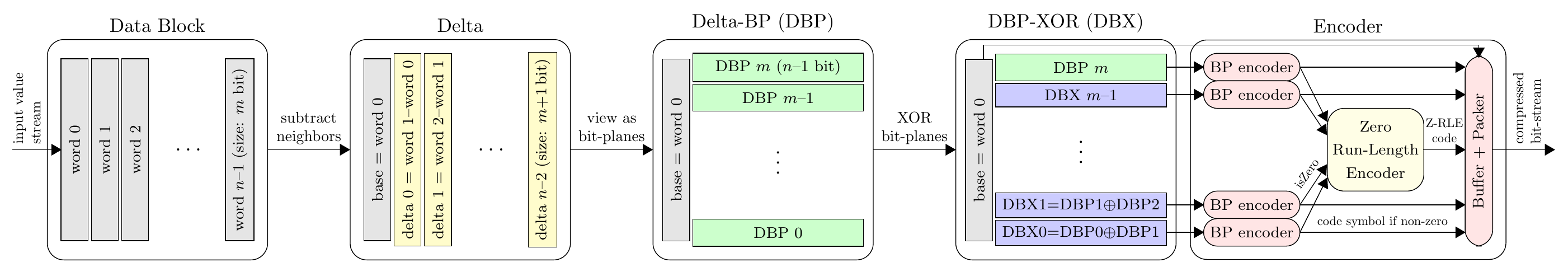}
	\caption{Overview of the processing steps to apply bit-plane compression.}
    \label{fig:BPCoverview}
\end{figure*}
\revOne{Besides applying these methods directly to the feature maps and using the already present sparsity to skip some of the compute operations, several methods such as DeltaRNNs \cite{Neil2016, Gao2018}, sigma-delta quantized networks \cite{Connor2017}, and CBinfer \cite{Cavigelli2017,Cavigelli2018} apply temporal differences to further sparsify the feature maps with the same intent. }

\subsection{\revOne{Model Compression}}
\revOne{More sophisticated compression methods have been proposed, particularly for compressing the model size. Most of them are very complex (silicon area) to implement in hardware.} One such method, deep compression \cite{Han2016}, combines pruning, trained clustering-based quantization, and Huffman coding. \novel{Most of these steps cannot be applied to the intermediate feature maps, which change for every inference as opposed to the weights which are static and can be optimized off-line. Furthermore, applying Huffman coding---while being optimal in terms of compression rate and given a specification of input symbols and their statistics---implies storing a very large dictionary: encoding a 16\,bit word requires a table with $2^{16}=65.5$k entries, but effectively multiple values would have to be encoded jointly in order to exploit their joint distribution (e.g., the smoothness), immediately increasing the dictionary size to $2^{2\cdot16}=4.29$G even for just two values.} 
Similar issues arise when using Lempel-Ziv-Welch (LZW) coding \cite{Welch1984,Ziv1978} as present in, e.g., the ZIP compression scheme, where the dictionary is encoded in the compressed data stream. This makes it unsuitable for a lightweight and energy-efficient VLSI implementation \cite{Lin2006,Zhou2016}. 

\subsection{\revOne{Compressed Representation Learning}}
Few more methods exist which change the CNN's structure in order to compress the weights \cite{Chen2015a,Agustsson2017} or the feature maps \cite{Gudovskiy2018,Wang2017a,Spallanzani2019}. However, they require altering the CNN's model and retraining it, and they introduce some accuracy loss. Furthermore, they can only be used to compress a few feature maps at specific points within the network and sometimes introduce additional compute effort, such as applying a Fourier transform to the feature maps. 

\subsection{\revOne{System-level Compression Methods}}
The most directly comparable approach, cDMA \cite{Rhu2018}, \revOne{relies on ZVC as introduced in NullHop \cite{Aimar2017}. }\novel{\revOne{However, }their target application differs in that their main goal is to allow faster temporary offloading of the feature maps from GPU to CPU memory through the PCIe bandwidth bottleneck during training, thereby enabling larger batch sizes and deeper and wider networks without sacrificing performance.} \revOne{They use ZVC in a configuration which takes a block of 32 activation values and generates a 32-bit mask where only the bits to the non-zero values are set. The non-zero values are stored and transferred after the masks. This provides the main advantage over Zero-RLE that the mask data has a fixed size and can easily be accessed even with non-linear access patterns, while this method provides small compression ratio advantages at the same time. Note that this can be seen as a special case of Zero-RLE with a maximum zero burst length of 1.}

For this work, we build on a method known in the area of texture compression for GPUs, \emph{bit-plane compression (BPC)} \cite{Kim2016}, fuse it with sparsity-focused compression methods, and evaluate the resulting compression algorithm \novel{on intermediate feature maps and gradient maps to show compression ratios of 5.1\x{} (8\,bit AlexNet), 4\x{} (VGG-16), 2.4\x{} (ResNet-34), 2.8\x{} (SqueezeNet), and 2.2\x{} (MobileNetV2).}

\section{Compression Algorithm}
\label{sec:algo}
\begin{figure}
	\includegraphics[width=\linewidth]{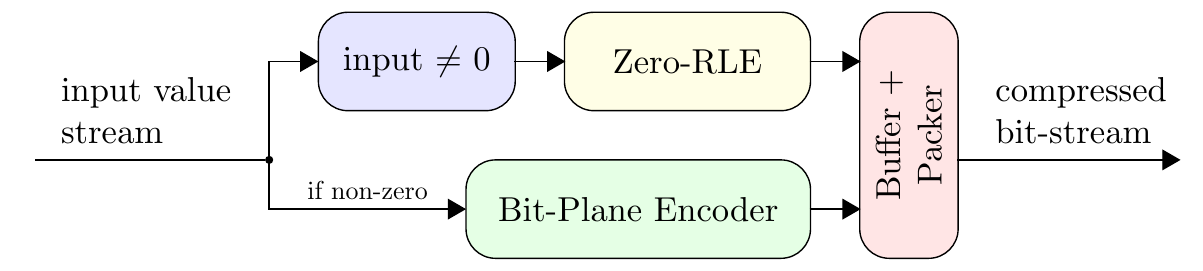}
	\caption{Top-level view of the proposed compression scheme. The bit-plane encoder is further illustrated in \figref{fig:BPCoverview}.}
    \label{fig:comprOverview}
\end{figure}

An overview of the proposed algorithm is shown in \figref{fig:comprOverview}. We motivate its structure based on the properties we have observed in feature maps (cf. \secref{sec:resultsDataProperties}): 
\begin{enumerate}
 \item Sparsity: The value stream is decomposed into a zero/non-zero stream on which we apply run-length encoding to compress the zero burst commonly occurring in the data. 
 \item Smoothness: Spatially neighboring values are typically highly correlated. We thus compress the non-zero values using bit-plane compression. The later compresses a fixed number of words $n$ jointly, and the resulting compressed bit-stream is injected immediately after at least $n$ non-zero values have been compressed.   
\end{enumerate}
The resulting algorithm can be viewed as an extension to bit-plane compression to exploit the sparsity present in most feature maps better. 

\subsection{Zero/Non-Zero Encoding with RLE}
\revOne{The run-length encoder produces a binary stream that specifies for each word of the input stream whether it is zero. Bursts of zeros are encoded by a '0' bit followed by a fixed number of bits describing the length of the burst. Non-zero inputs are not run-length encoded, and instead, each non-zero word is represented by a '1' bit.} If the length of a zero-burst exceeds the corresponding maximum burst length, the maximum is encoded, and the remaining bits are encoded independently, i.e., in the next code symbol. 

\subsection{Bit-Plane Compression}
An overview of the bit-plane compressor (BPC) used to compress the non-zero values is shown in \figref{fig:BPCoverview}. For BPC a set of $n$ words of $m$\,bit, a \emph{data block}, is compressed by first \revOne{computing} differences between every two consecutive words and storing the first word as the base. This exploits that neighboring values are often similar to reduce concentrates the distribution of the compressed values around zero. 

\begin{figure*}
    \begin{subfigure}{0.6\linewidth}
        \small\novel{
          \begin{tabular}{l@{}rl}
          	\toprule
          	DBX Pattern & Length [bit] & Code (binary) \\ 
          	\midrule
          	multi-all-0 DBX & $3+\ceil{\log_2 (m)}$ & 001 \& to\_bin(runLength-2) \\
          	all-0 DBX & 2 & 01 \\
          	all-1 DBX & 5 & 00000 \\
          	all-0 DBP & 5 & 00001 \\
          	2-consec 1s & $5+\ceil{\log_2(n-2)}$ & 00010 \& to\_bin(posOfFirstOne) \\
          	single-1 & $5+\ceil{\log_2(n-1)}$ & 00011 \& to\_bin(posOfOne) \\ 
          	uncompressed & $1+(n-1)$ & 1 \& to\_bin(DBX word) \\
          	\bottomrule
          \end{tabular}
          }
        \caption{\novel{Symbol Encoding Table}}
        \label{tbl:bitplaneSymbolEnc}
    \end{subfigure}
    \begin{subfigure}{0.4\linewidth}\centering
    	\includegraphics[width=\linewidth]{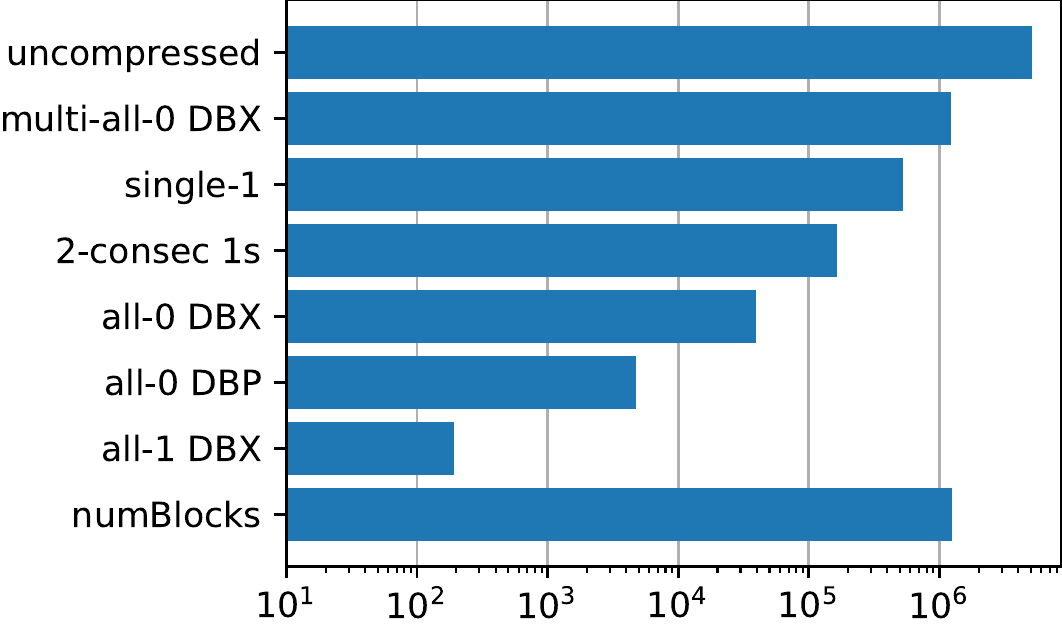}
    	\caption{\novel{Symbol Histogram and Block Count}}
        \label{fig:symbolCount}
    \end{subfigure}
    \caption{\novel{DBX and DBP symbol to code symbol mapping.}}
    \label{fig:codeSymbolCount}
\end{figure*}

The data items storing these differences are then viewed as $m+1$ bit-planes of $n-1$\,bit each (delta bit-planes, DBPs). Neighboring DBPs are XOR-ed, now called \emph{DBX}, and the DBP of the most significant bit is kept as the base-DBP. The results are fed into bit-plane encoders, which compress the DBX and DBP values to a bit-stream following \tblref{tbl:bitplaneSymbolEnc}. Most of these encodings are applied independently per DBX symbol. However, the first can be used to jointly encode multiple consecutive bit-planes at once, if they are all zero. This is where the correlation of neighboring values is best exploited. Note also the importance of the XOR-ing step in order to map two's complement negative values close to zero also to words consisting mostly of zero-bits. 

\novel{
As 1) the bit-plane encoding is a prefix code, 2) both the block size and word width are fixed, and 3) the representation of word\,0--$(n-1)$ as (base, DBP\,$m$, DBX\,0--$(m-1)$) is invertible, the resulting bit-stream of the base (word\,0) followed by all the encoded symbols can be decompressed into the original data block. 
}

\novel{
We have analyzed the code symbol distribution in \figref{fig:symbolCount} across all ResNet-34 feature maps with 8\,bit fixed-point quantization. Similar histograms are obtained for 16\,bit fixed-point and/or other networks. The 1.25\,M blocks result in 11.25\,M symbols of which 5.1\,M are uncompressed bit-planes, 1.2\,M are multi-all-0 DBX symbols encoding 5.4\,M all-zero bit-planes, 0.5\,M single-1 symbols, 0.2\,M symbols for bit-planes with two consecutive one-bits. 
}

\revOne{As we are processing a stream of data, transmitting the base can be omitted in favor of re-using the last word of the previous block. As the compression is loss-less, the last decoded word of the previous block is used as the base for decoding the next block. When starting to transmit a new stream of data, either base of the first block can be transmitted, or the base can be initialized to zero.
}

\subsection{\novel{Data Types}}\label{sec:dataTypes}
\novel{
The proposed compression method can be applied to integers of various word widths and for various block sizes. 
It also works with floating-point words, in which case the deltas do not need an additional bit and correspondingly there is one less DBP and DBX symbol (cf. \figref{fig:BPCoverview}). The floating-point subtraction is not exactly (bit-true) invertible. Hence a minimal and in practice negligible compression loss can be expected. 
Floating-point numbers are known to be notoriously hard to compress. While the DBX symbols corresponding to the fraction bits are almost equiprobable `1' or `0', those for the exponent and sign bits are often all-zero and thus compressible. 

Notably, this compression method is \revOne{capable of handling} variable-precision input data types very well. For example, 10\,bit values can be represented as 16\,bit values and fed into a bit-plane compressor for 16\,bit values. First, all the benefits coming from sparsity remain. Then, once a data block of such is converted into the DBX representation, there will now be 6 additional all-zero DBX symbols. These are then in the worst case encoded together into a single \emph{multi-all-0 DBX} symbol, adding a mere 7\,bit to the overall block's code stream. In the best case, the additional all-zero DBX symbols can be encoded into an existing adjacent symbol. 
Similarly, not only reduced bit-widths, but generally reduced value ranges will have a positive impact on the length of the compressed bitstream. This can be used to alter the trade-off between accuracy and energy-efficiency on-the-fly. 

While the focus here is on evaluating the compression rate on feature and gradient maps of CNNs, such a (de-)compressor will be beneficial for any smooth data (images/textures, audio data, spectrograms, biomedical signals, \dots) and/or sparse data (event streams, activity maps, \dots). 
}

\section{Hardware Architecture \& Implementation}
\label{sec:hw}
\novel{
The compression scheme's elements have been selected such that it is particularly suitable for a lightweight hardware implementation: no code-book needs to be stored, just a few data words need to be kept in memory. To verify this claim, we present a hardware architecture in this section from which we obtain implementation results. For both the compressor and decompressor, we chose to target a throughput of approximately 1 word per cycle. We have separate output data streams for the compressed zero/non-zero stream and the bit-plane compressed data, which could optionally be packed into a single compressed bitstream. In the following, we use a block size of 8 and 8\,bit fixed-point data words. 
\revOne{The implementation is available online\footnote{\url{https://github.com/pulp-platform/stream-ebpc}}}}.

\subsection{Compressor}
\begin{figure*}
	\includegraphics[width=\linewidth]{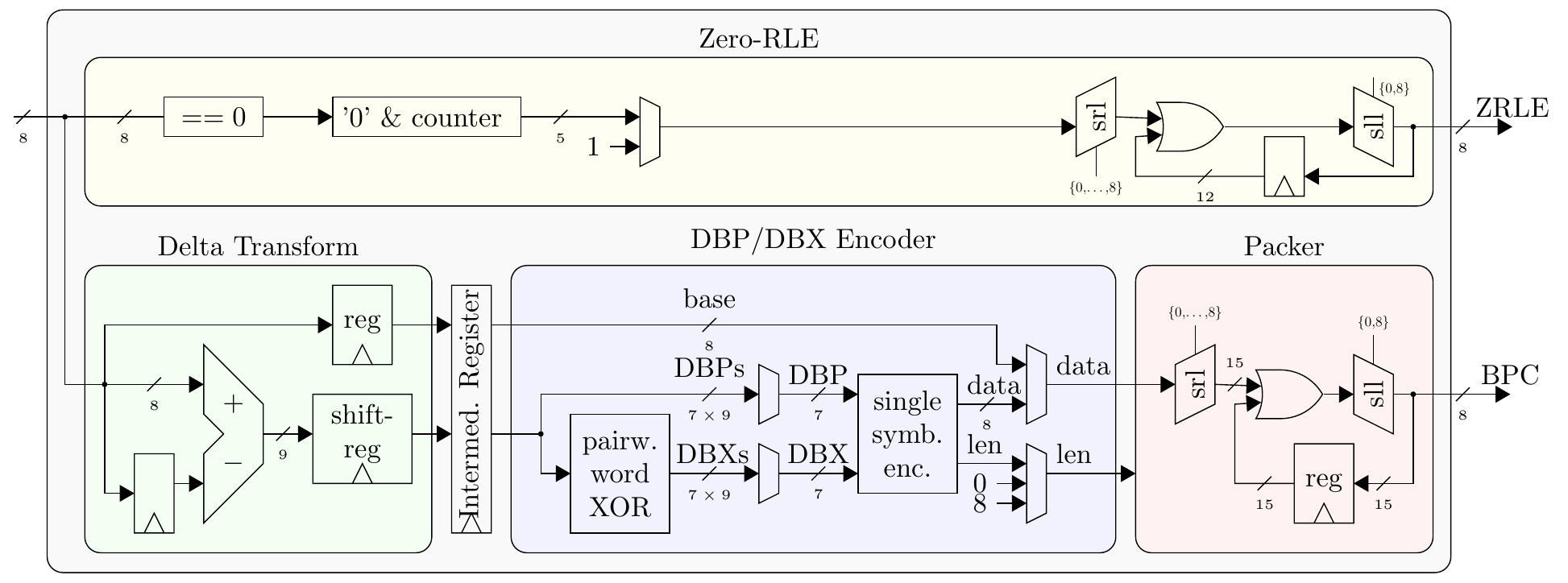}
	\caption{\novel{Block diagram of the full compressor without control logic as shown in \figref{fig:BPCoverview} for block size 8 and a word width of 8\,bit.}}
    \label{fig:bpEncBlockDiag}
\end{figure*}
\novel{
In \figref{fig:bpEncBlockDiag} we show the hardware architecture of the encoder. On top, we show the Zero-RLE compressor---a simple comparator to zero (an 8-input NOR block) followed by a counter and a multiplexer which selects a '1' in case of a non-zero or the zero count if one or more zeros have been received. Towards the end of the unit, variable-length symbols are packed into 8\,bit words for connection to a memory bus: a register is filled with shifted data until at least 8 bits have been collected, at which point an 8\,bit word is sent out, and the remaining bits in excess of 8 are shifted to the LSB side. At the same time, any non-zero values are processed by the \emph{Delta and DBP/DBX Transform} block. The first word is written to the base word register, all subsequent words of the block are each subtracted from their previous value, and these pushed into a shift-register which is read in parallel once a complete block has been aggregated (now interpretable as bit-planes) and the pair-wise XOR of the DBPs is taken to get the DBX symbols (for ease of implementation, the first DBP is XOR-ed with zero, i.e., directly taken as the DBX symbol). The entire block of symbols is then \revOne{stored in an intermediate buffer register}, such that new input words can be accepted while the bit-planes are iteratively encoded to allow an average throughput of up to 0.8 words/cycle. The data block is then read by the \emph{DBP/DBX encoder} to encode each bit-plane as a bit-vector and its length. The resulting variable-length data is then packed with a circuit similar to the packer in the \emph{Zero-RLE} block to produce fixed 8\,bit length words. 

Although the throughput of the bit-plane compression part of the circuit is limited to 0.8 word/cycle, this constitutes a worst-case scenario. When zero-values are encountered, the Zero-RLE block handles the workload while the processing of the non-zero words continues in parallel. This way, the compressor can be operated more closely to 1 word/cycle on average. 
}

\subsection{Decompressor}
\begin{figure*}
	\includegraphics[width=\linewidth]{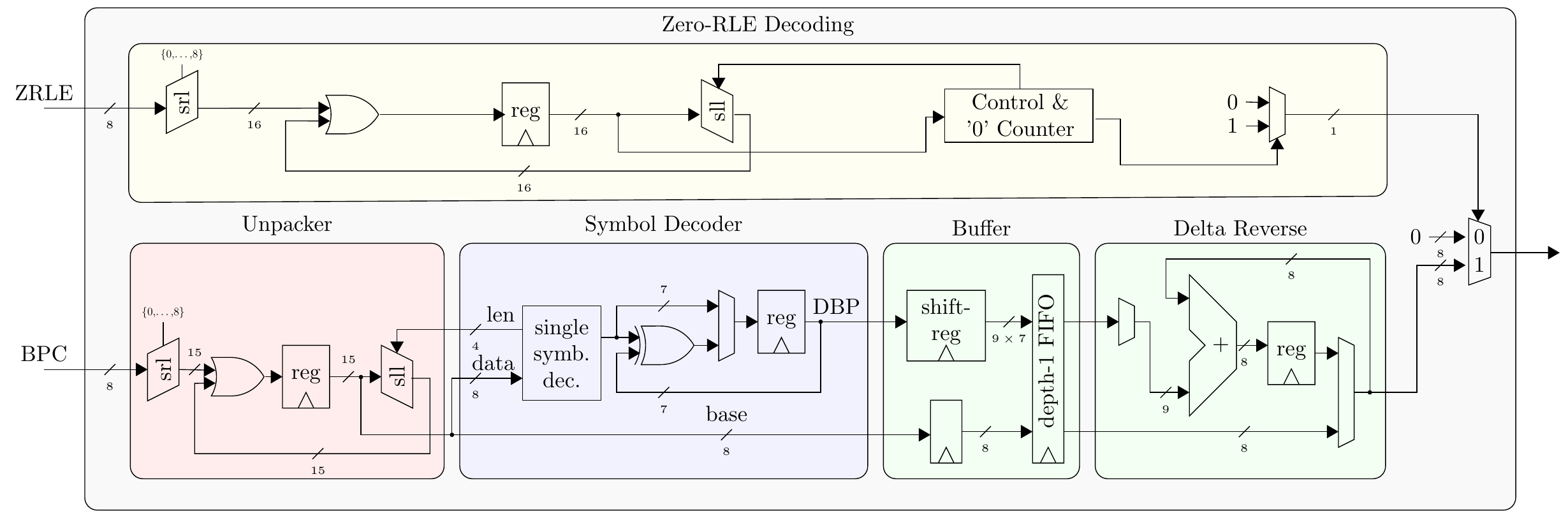}
	\caption{\novel{Block diagram of the full decompressor for block size 8 and a word width of 8\,bit without control logic.}}
    \label{fig:bpDecBlockDiag}
\end{figure*}
\novel{
The decompressor shown in \figref{fig:bpDecBlockDiag} reverts the steps of the encoder. After inverting the Zero-RLE encoding, the bit-plane compressed data stream is read in 8\,bit words, and unpacked into variable-length data chunks. The \emph{Unpacker} always provides 8 valid data bits to the \emph{Symbol Decoder}, which decodes the symbol into a DBP or DBX word and feeds the effective symbol length back to the \emph{Unpacker}. In case of a DBX word, it is XOR-ed with the previous DBP, such that DBP words are emitted to the \emph{Buffer} unit---or the base word is forwarded in case of the first 8 bits of the block. The \emph{Buffer} block aggregates the DBPs and the base word, buffering it for the \emph{Delta Reverse} unit, which iteratively accumulates the delta symbols and emits the decompressed words. The \emph{Buffer} unit with its built-in FIFO allows to unpack and decode data (10 cycles/block) while reverting the delta compression (8 cycles). 
}

\subsection{Implementation Results}
\novel{
\begin{table}
    \centering
    \caption{\novel{Area Cost of Silicon Implementation in UMC 65\,nm Technology for 600\,MHz Target Frequency}}
    \label{tbl:implArea}
\novel{\revOne{
\begin{threeparttable}
\begin{tabular}{l@{\hskip 1.5mm}l@{\hskip 1.5mm}r@{\hskip 1.5mm}rr@{\hskip 1.5mm}rr@{\hskip 1.5mm}r}
\toprule
    &  & \multicolumn{2}{c}{WW=8} & \multicolumn{2}{c}{WW=16} & \multicolumn{2}{c}{WW=32} \\
    &  & [\si{\micro\meter\squared}] & [GE]\tnote{a} & [\si{\micro\meter\squared}] & [GE] & [\si{\micro\meter\squared}] & [GE]  \\
\midrule
    \parbox[t]{2.5mm}{\multirow{5}{*}{\rotatebox[origin=c]{90}{Compressor}}}
    & Zero-RLE              &  685 &  476 & 1085 &  753 &  1806 & 1254 \\
    & Delta \& DBX Transf.  & 1059 &  735 & 1940 & 1347 &  3832 & 2661 \\
    & Depth-1 FIFO          &  871 &  605 & 1563 & 1085 &  3157 & 2192 \\
    & DBP/DBX Encoder       &  483 &  335 &  928 &  644 &  1750 & 1215 \\
    & Packer                &  432 &  300 &  808 &  561 &  1625 & 1128 \\
    \cmidrule{2-8}
    & Total                 & 4079 & 2833 & 6880 & 4778 & 12611 & 8792 \\ 
\midrule
    \parbox[t]{2.5mm}{\multirow{5}{*}{\rotatebox[origin=c]{90}{Decompressor\tnote{b}\,\,}}}
    & Unpacker              &  931 &  647 &  1997 & 1387  &  4369 & 3034 \\
    & Symbol Decoder        &  450 &  313 &  509 &  353 &   576 &  400 \\
    & Buffer                & 1597 & 1109 & 3056 & 2122 &  5940 & 4125 \\
    & Delta Reverse         &  359 &  249 &  620 & 431  &  1342 &  932  \\
    & Zero-RLE     &  812 &  564 & 1491 & 1035 & 3127  & 2172  \\
    \cmidrule{2-8}
    & Total        & 4330 & 3007 & 7986 & 5546 &  15160 & 10528 \\
\bottomrule
\end{tabular}
\begin{tablenotes}
\item[a] Gate equivalents (GEs): size expressed in terms of area of 2-input NAND gates. 1\,GE: 1.44\,$\mu$m${}^2$ (umc 65\,nm), 0.49\,$\mu$m${}^2$ (ST 28\,nm FDSOI), 0.20\,$\mu$m${}^2$ (GlobalFoundries 22\,nm).
\item[b] Without inverse Zero-RLE.
\item[c] The total area deviates from the sum of the blocks as the synthesizer performs optimizations across the blocks \revOne{and because the top-level blocks contain some logic as well}.
\end{tablenotes}
\end{threeparttable}
}
}
\end{table}
\revOne{
We have implemented the described architecture for a UMC 65\,nm low-leakage process and synthesized the design using the Synopsys Design Compiler 2018.06 and UMC 65\,nm low-leakage cell libraries in the typical, 25\si{\celsius} corner. We report the area and a per-unit breakdown for a block size of 8 (i.e., the optimal case, cf. \secref{sec:bpcBlockSize}) and 8\,bit, 16\,bit and 32\,bit words and a target frequency of 600\,MHz in \tblref{tbl:implArea}.  For most inference applications, 8\,bit feature maps are sufficient, such that the compressor and decompressor fit onto a mere 4000\,\si{\micro\meter\squared} and 4330\,\si{\micro\meter\squared}, respectively. For comparison, the area of both together is a bit less than 7$\times$ that of an 8-bit multiply-add unit\footnote{\revOne{Using the Synposys DesignWare MAC unit.}} at 600\,MHz, which requires 842\,\si{\micro\meter\squared}. }

Synthesis of the circuit for lower frequencies does not reduce area while at 1\,GHz and 1.5\,GHz the area of the compressor grows to 4337\,\si{\micro\meter\squared} and 5180\,\si{\micro\meter\squared}, respectively. 
For higher frequencies, timing closure could not be attained, with the longest path passing from the DBX multiplexer's control input in the \emph{DBP/DBX Encoder} to the register in the \emph{Packer}. 

Directly scaling up the word size from 8\,bit to 16\,bit increases the area of the compressor by 70\% and 85\% for the decompressor. Increasing it further from 16\,bit to 32\,bit, adds another 83\% or 108\%, respectively. Increasing the word width does not have any effect on the \emph{DBP/DBX Encoder}, as it works on bit-planes, but it requires more iterations. It might thus be considered using multiple \emph{DBP/DBX Encoder} units not to bottleneck the throughput and thereby also increasing the size of the \emph{Packer} and \emph{Unpacker} to be able to take data from all encoder and feed all the decoders. The size of the \emph{Packer} and \emph{Unpacker} increases as well with the word width as the register size grows, and so does the number of multiplexers in the barrel shifters. 

Scaling up the throughput can be achieved by doubling the capacity of each unit, reading two words into the compressor, computing the differences both in the same cycle, and increasing the size of the input port of the shift register. Similarly, multiple encoders can be used to compress two bit-planes per cycle. This will only have a limited impact on the area in this part of the compressor, which is mostly defined by the size of the shift-register and the FIFO, which do not need to grow. The main impact will be visible within the \emph{Packer} and later the \emph{Unpacker} units, where the barrel shifters have to take twice as wide words and shift twice as far when doubling the throughput, and hence grows quadratically---a problem inherent to packing data of any variable symbol-size compressors. For the decompressor, similarly, there can be multiple symbol decoders and the \emph{Delta Reverse} unit can be modified to process two words per cycle. Overall, increasing the throughput this way can be expected to scale below linear in area for processing few words in parallel, but once reaching close to full parallelization (i.e., 8 for block size 8 and word width 8\,bit), the size of the \emph{Packer} and \emph{Unpacker} will take up most of the circuit's size. However, the throughput can be scaled with linear area cost by instantiating multiple complete (de-)compressors to work on individual feature maps in parallel or on separate spatial tiles of the feature maps. 
}

\subsection{System Integration}
\novel{
The presented compression scheme can be used to reduce the energy spent on interfaces to external DRAM, on inter-chip or back-plane communication---the corresponding standards specify very efficient power-down modes \cite{lpddr4,gddr5}---and to reduce the required bandwidth of such interfaces, lowering the cost of packaging, circuit boards, and additional on-chip circuits (e.g., PLLs, on-chip termination, etc.) \cite{lpddr4,gddr5}. 

Given the limited size, it can also be used to reduce the size of on-chip data transfers, e.g., from large background L2 memories in large DNN inference chips that try to fit all data on chip, such as the one Tesla has presented for its next generation of self-driving cars or the hardware by Graphcore \cite{Toon2017}.

\paragraph{Streaming HW Accelerators} 
The (de-)compressor could be integrated with an accelerator such as YodaNN \cite{Andri2018a} which reaches a state-of-the-art core energy efficiency of 60\,TOp/s/W for binary-weight DNNs. For the specific case of YodaNN, however, taking I/O energy cost into accounts adds 15.28\,mW to the core's 0.26\,mW, bottlenecking the efficiency to 1 TOp/s/W. A drop-in addition of 8 compressor and decompressor units---YodaNN works on 8 feature maps at the input and output in parallel---would reduce the I/O cost and directly increase its energy efficiency at system level by 2--4\x{} (cf. \secref{sec:totalComprFactor}) while adding only 0.05\,mm${}^2$ (6\%) to the 0.86\,mm${}^2$ of core area. 

\paragraph{HW Accelerator with Feature Maps On-Chip} 
Another application scenario would be with Hyperdrive \cite{Andri2019} and large industrial chips such as Tesla's platform for its next generation of cars, which store the feature maps on chip. Memory inherently takes up a large share of such a design, for the case of Hyperdrive, specifically 65\%. With a compression scheme providing a reliable compression ratio across different input images and for all layer pairs (in a ping-pong buffering scheme), we can reduce the memory size by around 2\x{} (cf. \secref{sec:perLayerComprRatio}), to saving almost as much silicon area. 

\paragraph{Integration into a Heterogeneous Many-Core Accelerator}
A further use-case is the integration into a heterogeneous accelerator with multiple cores and/or accelerators working from a local scratchpad memory, where data is prefetched from a different level in the memory hierarchy, e.g., in the \emph{GAP-8} SoC \cite{Flamand2018} which can be used for DNN-based autonomous navigation of nano-drones \cite{Palossi2018}, 8 cores and a CNN hardware accelerator which share a 64\,kB L1 scratchpad memory which is loaded with data from the 512\,kB L2 memory using a DMA controller. In such systems, SRAM memory accesses and data movement across interconnects can make up for a significant share of the overall power, and generally memory space is a scarce resource. Integration of the proposed (de-)compressor into the DMA would improve both aspects jointly in such a system. 
}

\section{Results}
\label{sec:results}
\subsection{Experimental Setup}
Where not otherwise stated, we perform our experiments on AlexNet and are using images from the ILSVRC validation set. \novel{The models we used were pre-trained and downloaded from the PyTorch/Torchvision data repository wherever possible, and an identical preprocessing was applied to the data. \revOne{As our method is lossless, we reach  (top-1/top5) error rates of 43.45\%/20.91\% with AlexNet, 28.41\%/9.62\% with VGG-16, 26.70\%/8.58\% with ResNet-34,  41.81\%/19.38\% with SqueezeNet, and 28.12\%/9.71\% with MobileNetV2}. For the gradient analyses, we self-trained the same networks\footnote{Using code available at \url{https://github.com/spallanzanimatteo/QuantLab}}. Some of the experiments are performed with fixed-point data types (default: 16-bit fixed-point). The feature maps were normalized to span 80\% of the full range before applying uniform quantization in order to use the full value range up to a safety margin to prevent overflows. All the feature maps were extracted after the ReLU (or ReLU6) activations. }\revOne{The code to reproduce these experiments is available online\footnote{Code: \url{https://github.com/lukasc-ch/ExtendedBitPlaneCompression}}. }

\subsection{Sparsity, Activation Histogram \& Data Layout}
\label{sec:resultsDataProperties}
\begin{figure*}
	\includegraphics[width=\linewidth]{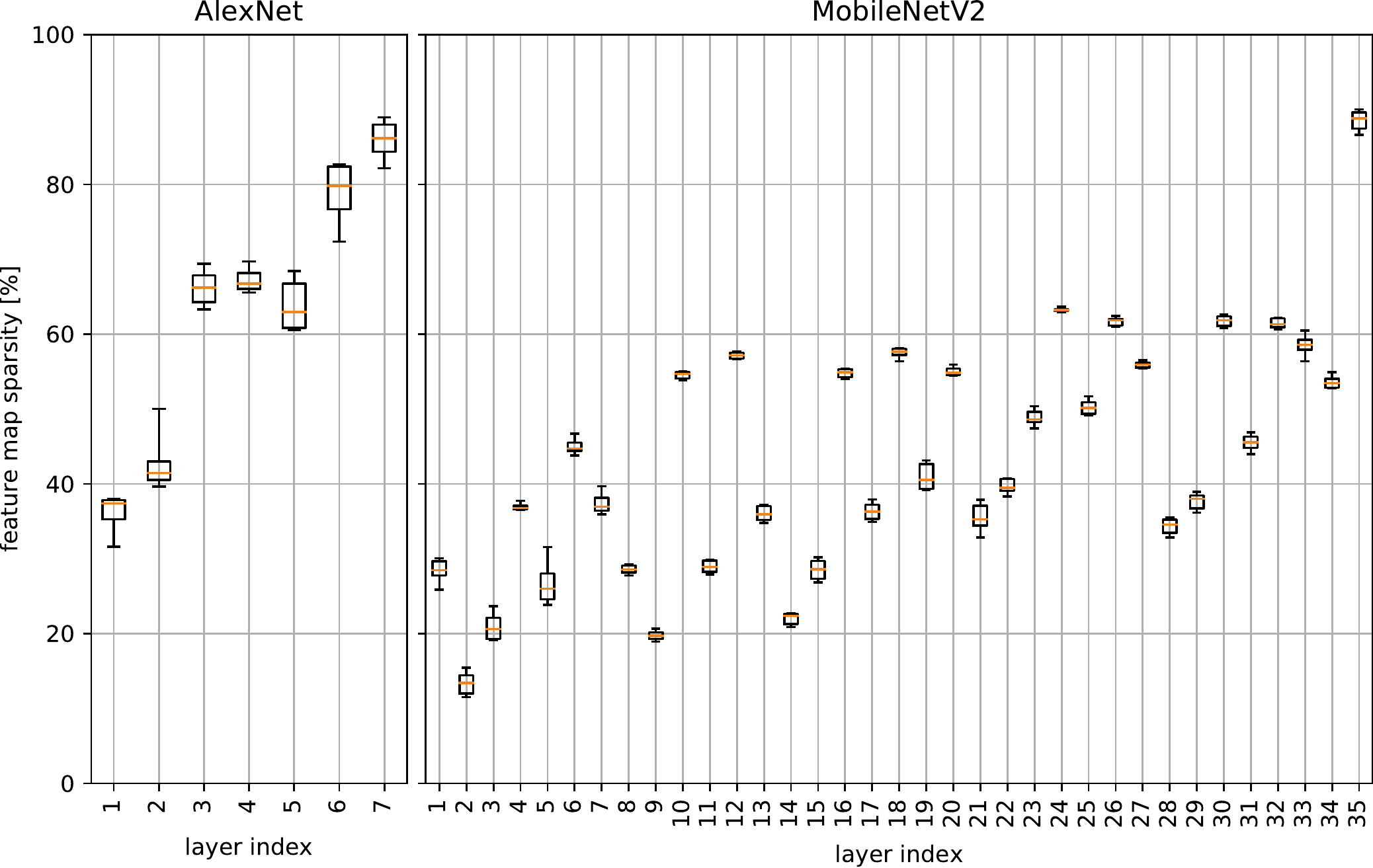}
	\caption{\novel{Feature map sparsity after activation for each layer of the fully trained network with information on the distribution across 250 frames of the validation set (median, 1st and 3rd quartile, 1st and 99th percentile).}}
	\label{fig:sparsity}
\end{figure*}

Neural networks are known to have sparse feature maps after applying a ReLU activation layer, which can be done on-the-fly after the convolution layer and possibly batch normalization. However, it varies significantly for different layers within the network as well as for different CNNs. \novel{Sparsity is a key aspect when compressing feature maps, and we analyze it quantitatively with statistics collected across 250 random ILSVRC'12 validation images and for each layer of AlexNet as well as the more modern and size-optimized MobileNeV2 in \figref{fig:sparsity}. For AlexNet, we can clearly see the increase in sparsity from earlier to later layers. For MobileNetV2, multiple effects are overlying. Overall, the feature maps later in the network are more sparse, and generally this is correlated with the number of feature maps (also in AlexNet). Feature maps following expanding 1\x1 convolutions (e.g., 15, 17, 19, 21) generally show lower sparsity (25--40\%) than after the depth-wise separable 3\x3 convolutions (e.g., 16, 18, 20, 22; sparsity 50--65\%), where for the latter there are exceptions (e.g., 8, 14, 28) when these convolutions were strided (sparsity 20--35\%). This aligns with intuition as the 1\x1 layers combine feature maps to be filtered later, and the depth-wise 3\x3 convolution layers literally perform the filtering. 

Besides the average sparsity, its probability distribution across different frames becomes relevant in case guarantees have to be provided either due to real-time requirements in case of a bandwidth-limited hardware accelerator or due to size limits of the memory in which the feature maps are stored (e.g., on-chip SRAM). The whiskers in the box plot mark the 1st and 99th percentile, clearly showing how narrow the distribution of the sparsity is and that we thus can expect a very stable compression rate. 
}

\begin{figure*}
    \begin{subfigure}{0.48\linewidth}
        \centering
	    \includegraphics[width=\linewidth]{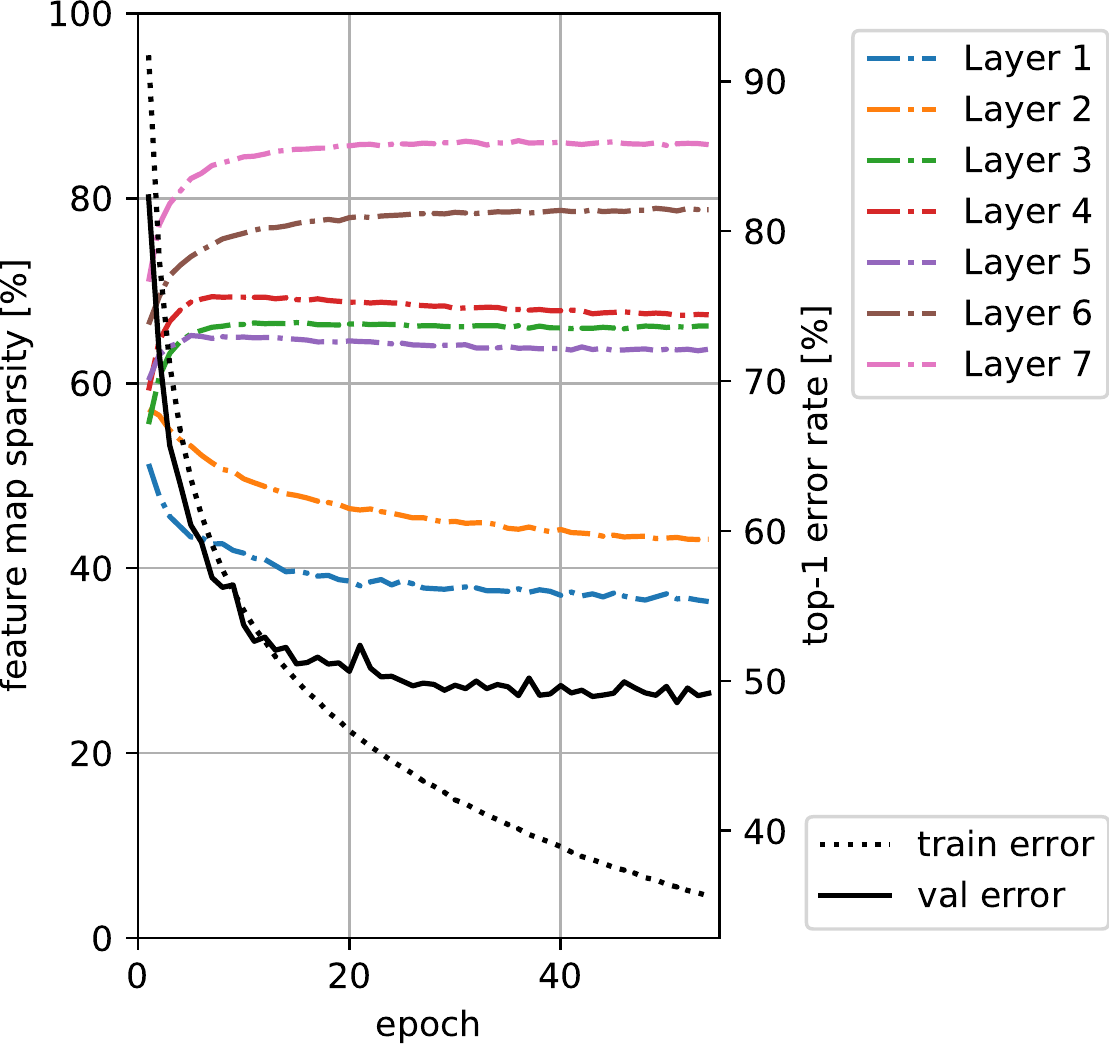}
	    \caption{AlexNet}
    \end{subfigure}
    \hfill
    \begin{subfigure}{0.48\linewidth}
        \centering
	    \includegraphics[width=\linewidth]{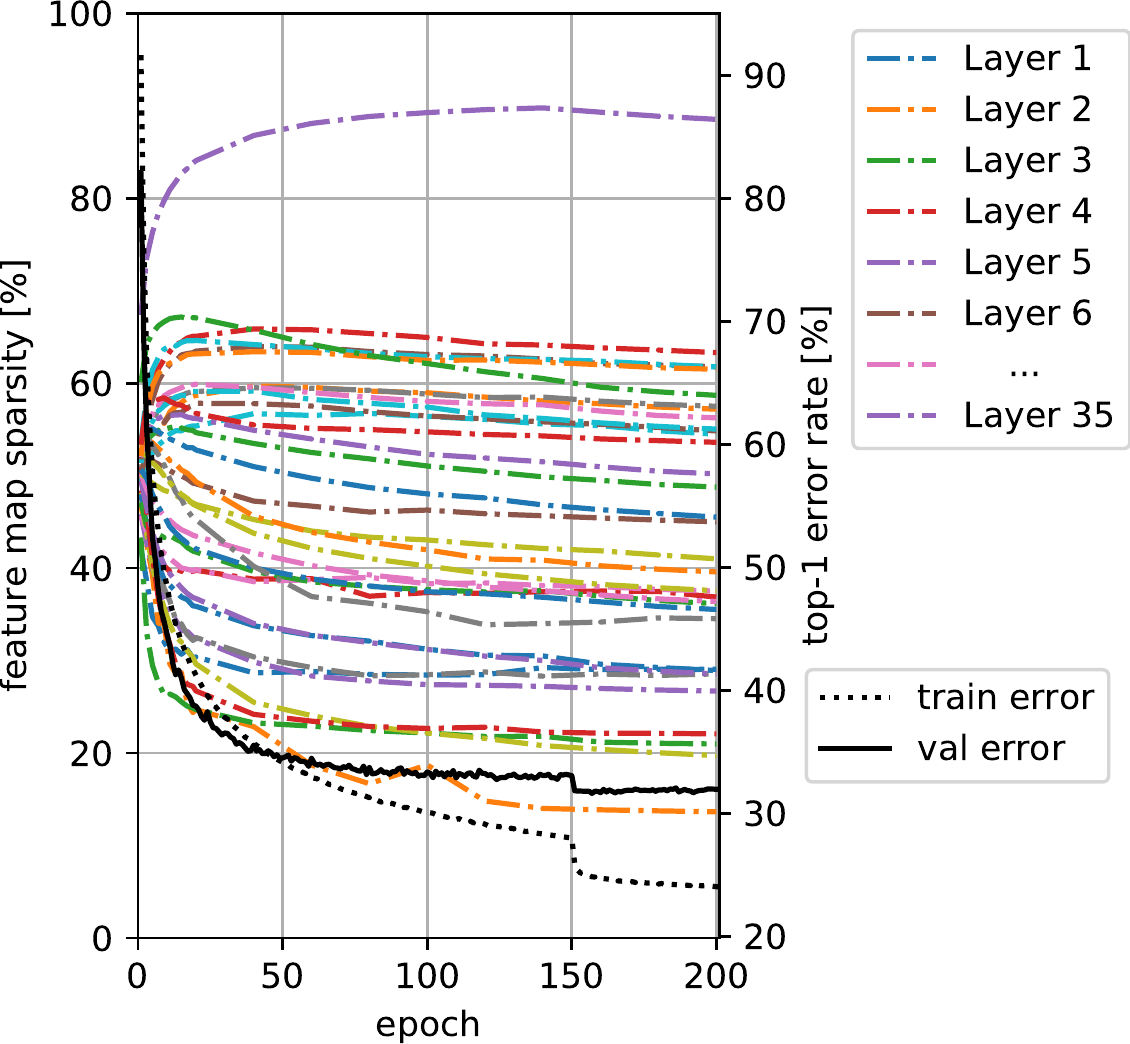}
	    \caption{MobileNetV2}
    \end{subfigure}
	\caption{\novel{Feature map sparsity after activation by epoch for each layer. Note that the gradients are as sparse as the activations, since a zero activation results in a zero gradient.}}
	\label{fig:sparsityByEpoch}
\end{figure*}
\novel{
We consider compressing not only the feature maps but also the gradient maps for specialized training hardware, thus raising the question of how sparsity evolves over as training progresses. The gradient maps are generally identically sparse as the corresponding feature maps, as ReLU activations pass a zero-gradient wherever the outgoing feature map value was zero. In \figref{fig:sparsityByEpoch}, we can observe how the various layers are starting from all-50\% after random initialization and with a few epochs settle close to their final level. In both networks, this is the case after around 15\% of the epochs required for full convergence. 
}

\begin{figure*}
	\centering
	\includegraphics[width=\linewidth]{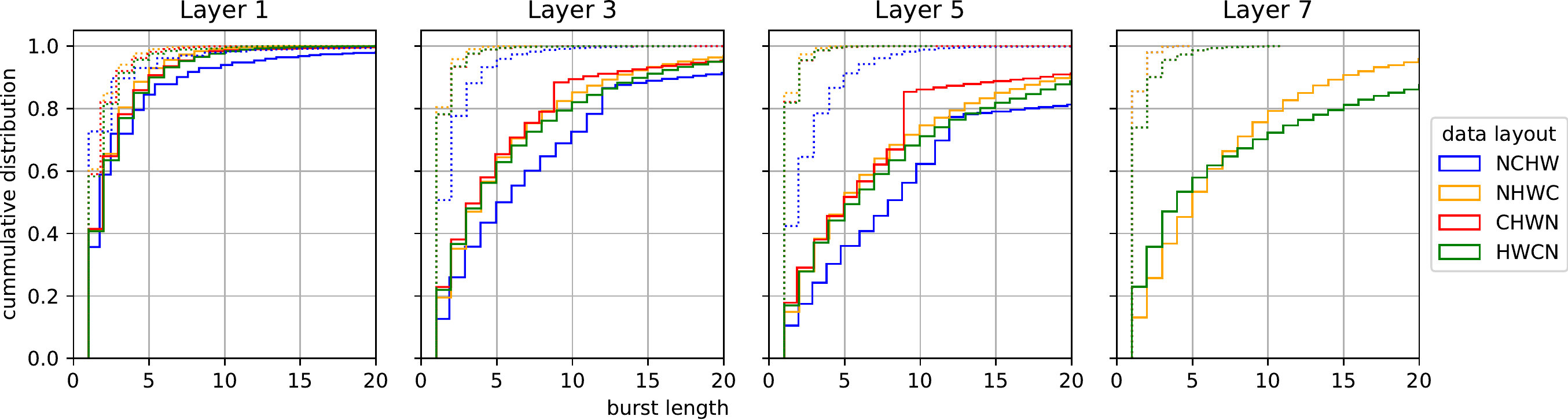}
	\caption{Cumulative probability distribution of zero (solid) and non-zero (dotted) burst lengths of activated feature maps in AlexNet.}
	\label{fig:zeroBurstLens}
\end{figure*}

The sparse values are not independently distributed but rather occur in bursts when the 4D data tensor is laid out in one of the obvious formats. The most commonly used formats are NCHW and NHWC, which are those supported by most frameworks and the widely used Nvidia cuDNN backend. NCHW is the preferred format for cuDNN and the default memory layout and means that neighboring values in the horizontal direction are stored next to each other in memory before the vertical, channel, and batch dimensions. NHWC is the default format of TensorFlow and has long before been used in computer vision and has the advantage of simple non-strided computation of inner products in channel (i.e., feature map) dimension. Further reasonable options that we include in our analysis are CHWN and HWCN, although most use-cases with hardware acceleration are targeting real-time low-latency inference and are thus operating with a batch size of 1. We analyze the distribution of the length of zero bursts for these four data layouts at various depths within the network in \figref{fig:zeroBurstLens}. 

The results clearly show that having the spatial dimensions (H, W) next to each other in the data stream provides the longest zero bursts (lowest cumulative distribution curve) and thus a better compressibility than the other formats. This is also aligned with intuition: feature maps values mark the presence of certain features and are expected to be smooth \cite{Mahmoud2018}. Inspection the feature maps of CNNs is commonly known to show that they behave like 'heat maps' marking the presence of certain geometric features nearby. Based on these results, we perform all the following evaluations based on the NCHW data layout. 

Note also that the burst length of non-zero values is mostly very short, such that there is limited gain in applying RLE also for the one-bits. \revOne{Specifically, the probability of the burst length of non-zero values being at most 3 is 90\%, 88\%, 80\%, and 95\% for layer 1, 3, 5, and 7, respectively. Using run-length encoding for such mostly very short burst introduce a significant overhead which is not outweighed by the benefits. }

\begin{figure*}
	\centering
	\novel{
    \begin{subfigure}[b]{0.48\linewidth}
        \centering
	    \includegraphics[width=\linewidth]{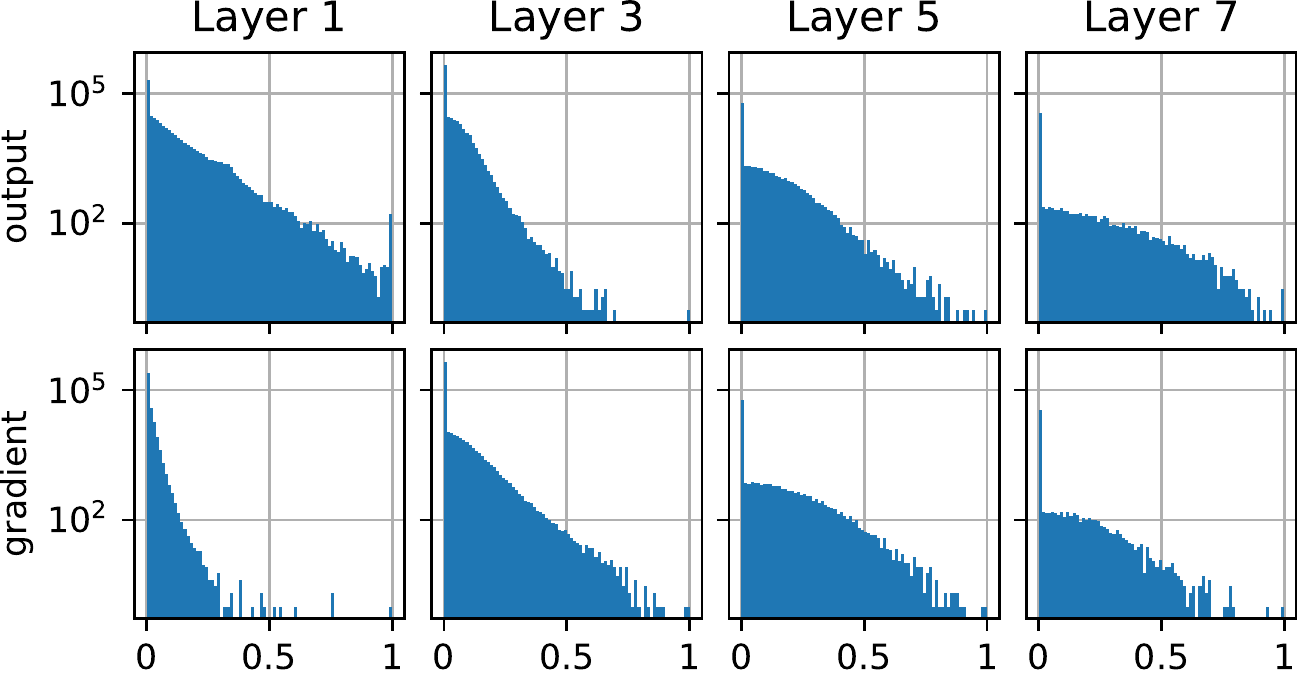}
	    \caption{AlexNet}
    \end{subfigure}
    \hfill
    \begin{subfigure}[b]{0.48\linewidth}
        \centering
	    \includegraphics[width=\linewidth]{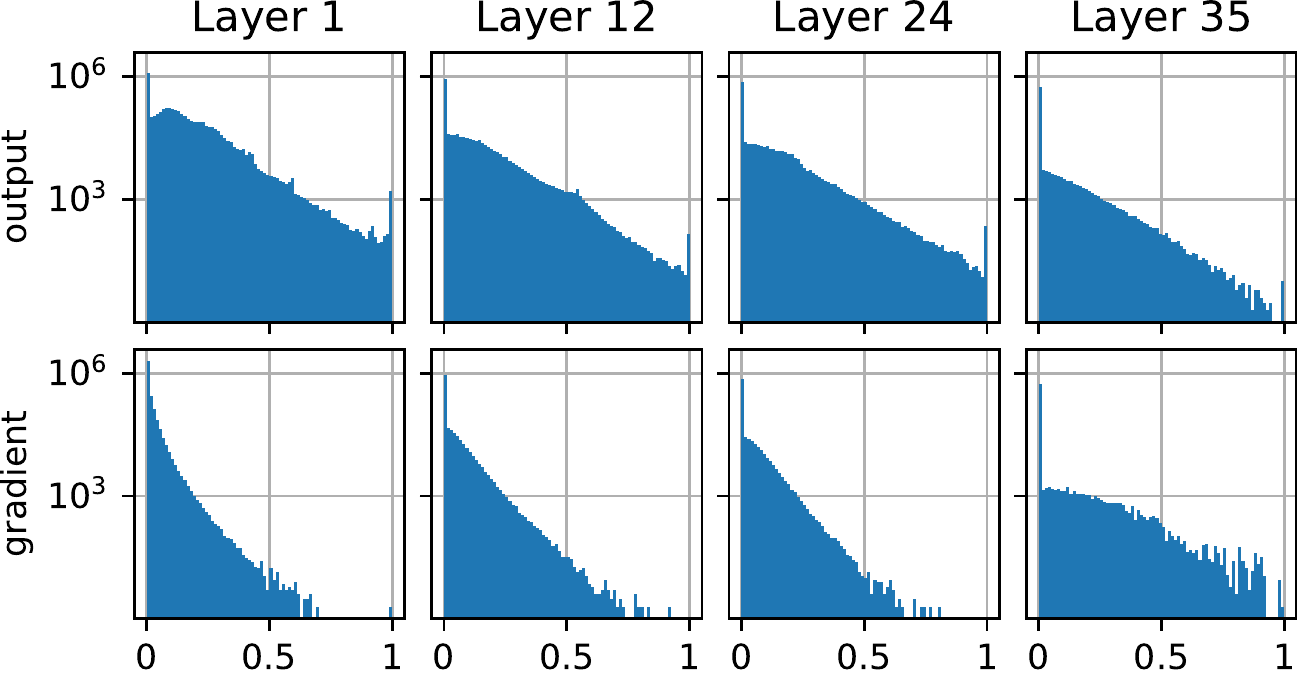}
	    \caption{MobileNetV2}
    \end{subfigure}
	\caption{\novel{Histogram of activation (top) and gradient (bottom) values at various depths within the network. Note the logarithmic vertical axis.}}
	\label{fig:hist}
	}
\end{figure*}

To compress further beyond exploiting the sparsity, the data has to remain compressible. This is definitely the case as can be seen when looking at histograms of the activation distributions as shown for some sample layers of AlexNet and MobileNetV2 in \figref{fig:hist} and a strong indication that additional compression of the non-zero data is possible. 

\subsection{Selecting Parameters}
The proposed method has two parameters: the maximum length of a zero sequence that can be encoded with a single code symbol of the Zero-RLE, and the BPC block size ($n$, number of non-zero words encoded jointly). 

\paragraph*{Max. Zero Burst Length}
We first analyze the effect of varying the maximum zero burst length for Zero-RLE on the compression ratio without for various data word widths in \tblref{tbl:zeroRLEburstLen}. 
\begin{table}
	\centering
	\caption{Compression Ratio Using ZVC And Zero-RLE for Various Maximum Zero Burst Lengths}
	\label{tbl:zeroRLEburstLen}
	\begin{tabular}{r|r|rrrrrr}
	\toprule
		\multirow{2}{*}{word width} & \multirow{2}{*}{ZVC} & \multicolumn{6}{c}{Zero-RLE max. zero burst length} \\
	    &    &   $2^1$ &   $2^2$ &   $2^3$ &   $2^4$ &   $2^5$ &   $2^6$ \\
	\midrule
	  8 &  2.52 &  2.48 &  2.56 &  2.62 &  \textbf{2.63} &  2.59 &  2.53 \\
	 16 &  3.00 &  2.96 &  3.02 &  3.06 &  \textbf{3.07} &  3.04 &  3.00 \\
	 32 &  3.30 &  3.28 &  3.32 &  3.34 &  \textbf{3.35} &  3.33 &  3.31 \\
	\bottomrule
	\end{tabular}
\end{table}
The optimal value is arguably the same for our proposed method, since a constant offset in compressing the non-zero values does not affect the optimal choice of this parameter (just like the word width has no effect on it). The results also serve as a baseline for Zero-RLE and ZVC. It is worth noting that ZVC corresponds to Zero-RLE with a max. burst length of 1, yet breaks the trend shown in \tblref{tbl:zeroRLEburstLen}. This is due to an inefficiency of Zero-RLE in this corner: for a zero burst length of 1, ZVC requires 1\,bit whereas Zero-RLE with a max. burst length of 2 takes 2\,bit. For a zero burst of length 2, ZVC encode 2 symbols of 1\,bit each and Zero-RLE takes 2\,bit as well. ZVC thus always performs at least as well for such a short max. burst length. 

\begin{figure}
	\centering
	\includegraphics[width=\linewidth]{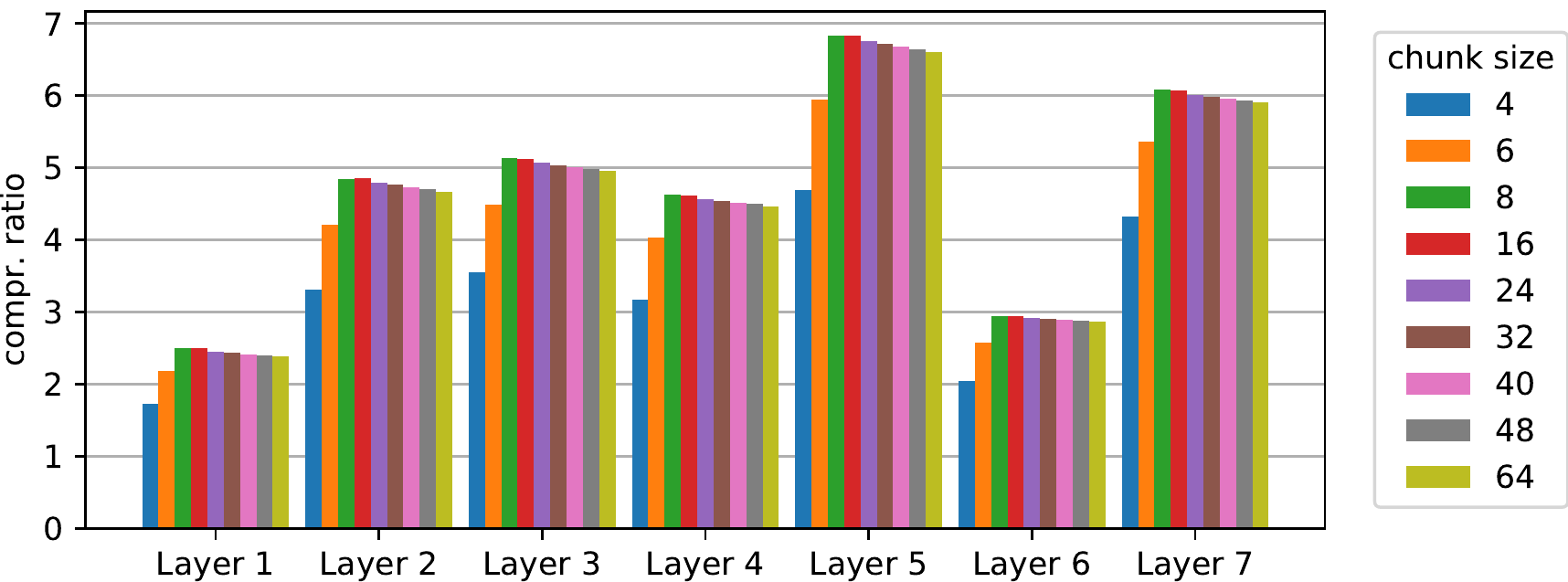}
	\caption{\novel{Analysis of the compression ratios of various layers' outputs for several BPC block sizes and with 16-bit fixed-point values.}}
	\label{fig:chunkSizeEval}
\end{figure}

\paragraph*{BPC Block Size} \label{sec:bpcBlockSize}
We analyze the effect of the BPC block size parameter in \figref{fig:chunkSizeEval} at various depths within the network. The best compression ratio is achieved with a block size of 16 across all the layers. A block size of 8 might also be considered to minimize the resources of the (de-)compression hardware block at a small drop in the compression ratio. 

\subsection{Total Compression Factor} \label{sec:totalComprFactor}
\novel{
\begin{figure*}
	\centering
	\includegraphics[width=\linewidth]{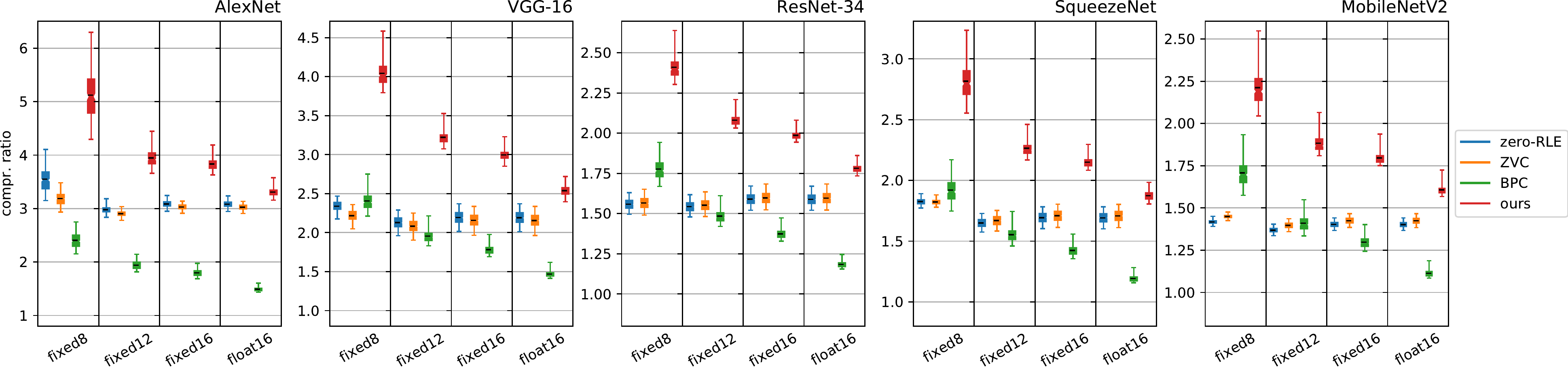}
	\caption{
    \novel{Compression ratio of all feature maps of AlexNet, VGG-16, ResNet-34, SqueezeNet, and MobileNetV2 for various compression methods and data types.}}
	\label{fig:totalComprRate}
\end{figure*}
We analyze the total compression factor of all feature maps of AlexNet, VGG-16, ResNet-34, SqueezeNet, and MobileNetV2 in \figref{fig:totalComprRate}. For AlexNet, we can notice the high compression ratio of around 3\x{} already introduced by Zero-RLE and ZVC and that it is very similar for all data types. We further see that pure BPC is not suitable since it introduces too much overhead when encoding only zero-values. For ResNet-34, SqueezeNet, and MobileNetV2, the gains by exploiting only the sparsity is significantly lower at around 1.55\x{}, 1.7\x{} and 1.4\x{}. The proposed method outperforms previous approaches clearly and particularly for 8\,bit fixed-point values commonly used in today's inference accelerators. There we observe compression ratios of 5\x{} (AlexNet), 4\x{} (VGG-16), 2.4\x{} (ResNet-34), 2.8\x{} (SqueezeNet), and 2.2\x{} (MobileNetV2). 

When moving from 8\,bit fixed-point values to 12 and 16\,bit and ultimately to 16\,bit floating point, the compression ratio of the methods based on sparsity only (zero-RLE, ZVC) do not change significantly. This is in line with expectations since the zero-values only make a negligible contribution to the final data size with zero-RLE and do not have any effect with ZVC. Contrary to this, BPC is very effective for small word widths but loses its benefits as word widths increase. Our proposed method combines the best of the two worlds, starting with a very high compression ratio and slowly converging to ZVC and zero-RLE as the word width increases. 
The gains for 8-bit fixed-point data are significantly higher than for other data formats. Most input data---also CNN feature maps---carry the most important information is in the more significant bits and in case of floats in the exponent. The less significant bits appear mostly as noise to the encoder and cannot be compressed without accuracy loss, such that this behavior---a lower compression ratio for wider word widths---is expected. 
}

\subsection{Per-Layer Compression Ratio} \label{sec:perLayerComprRatio}
\begin{figure*}
	\centering
	\includegraphics[width=\linewidth]{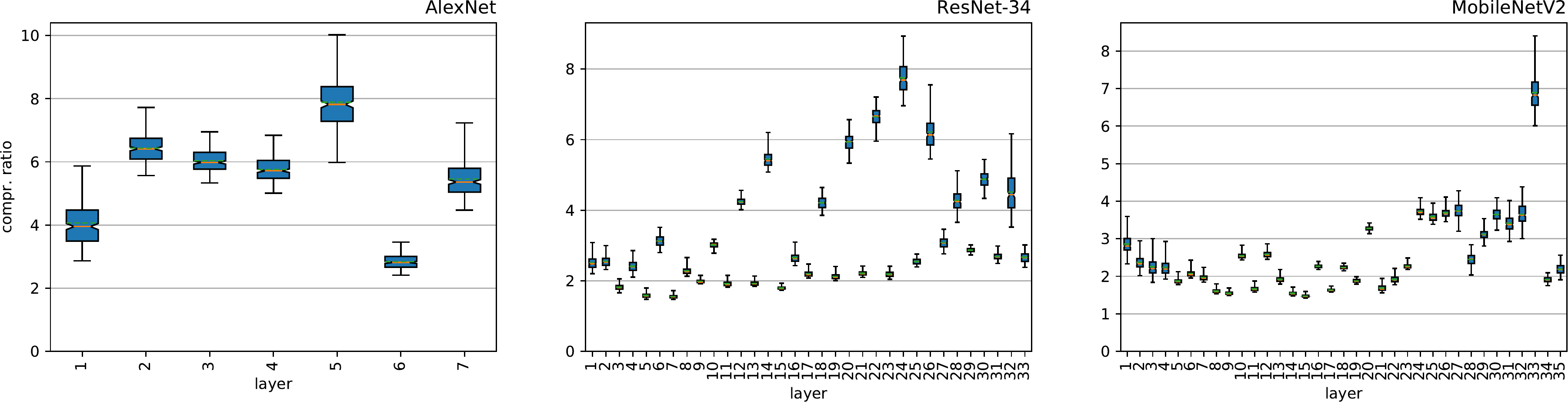}
	\caption{
    \novel{Per-layer compression ratio for AlexNet, ResNet-34, and MobileNetV2.}}
	\label{fig:perLayerComprRate}
\end{figure*}
\novel{
As already expected from the results on sparsity, the compression ratio is fairly stable across multiple frames. Specifically, the 1st percentile of compression ratios only lies around 20\% below the average case for all the networks. Further results on a per-layer basis for AlexNet, ResNet-34, and MobileNetV2 is shown in \figref{fig:perLayerComprRate}. While there is significant variability between the layers, the 1\% farthest outliers towards lower compression ratios can be found in AlexNet's first layer at a drop of around 25\% from the average ratio. For ResNet-34 and MobileNetV2, even the worst-case variations remain within less than 5\% deviation from the mean. This allows us to scale down the available bandwidth and/or the corresponding memory size with only a minimal risk of failure. Furthermore, the remaining risk can be further reduced when processing video streams in real-world applications, where the numerical precision could be scaled down to graciously as described in \secref{sec:dataTypes}, thus allowing to accept graceful accuracy degradation in exchange for a smaller size of the compressed bitstream, thereby mitigating potentially catastrophic failure. 
}

\begin{figure*}
	\centering
	\includegraphics[width=\linewidth]{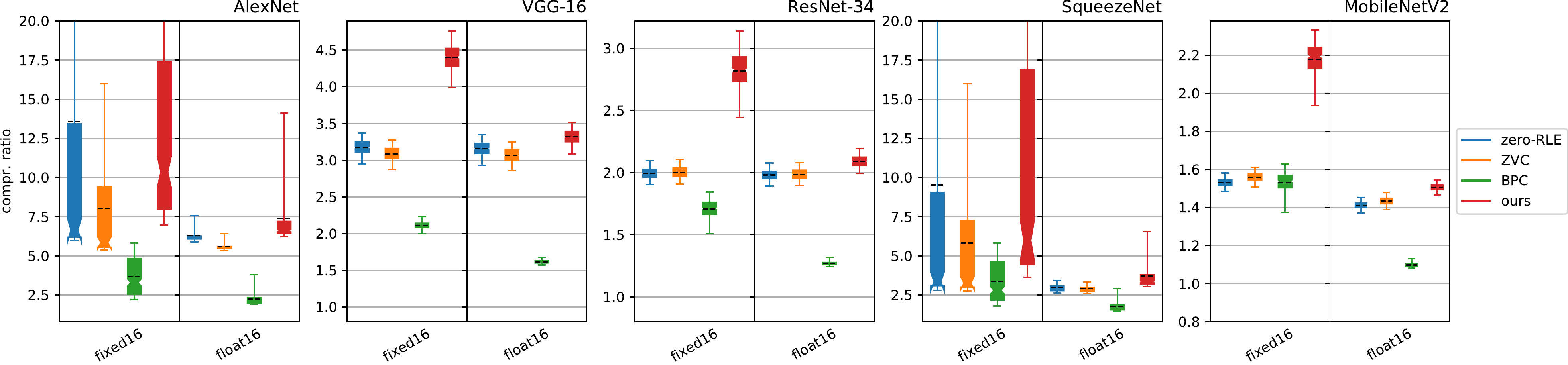}
	\caption{
    \novel{Compression ratio of all the gradient maps for various networks and several compression methods and data types.}}
	\label{fig:totalComprRateGrads}
\end{figure*}
\novel{
For applications in on-device learning as well as to further boost the throughput of thermally or I/O-limited training accelerators in computing clusters, we have further investigated the compressibility of the gradient maps (cf. \figref{fig:totalComprRateGrads}). Despite the higher precision data types as required for the gradients, high compression rates can be achieved, mostly higher or on par with those of the feature maps. 
}

\section{Conclusion}
\label{sec:conclusion}
\novel{
We have presented and evaluated a novel compression method for CNN feature maps. The proposed algorithm achieves an average compression ratio of 5.1\x{} on AlexNet (+46\% over previous methods), 4\x{} on VGG-16 (+67\%), 2.4\x{} on ResNet-34 (+33\%), 2.8\x{} on SqueezeNet (+51\%), and 2.2\x{} on MobileNetV2 (+30\%) for 8\,bit data, and thus clearly outperforms state-of-the-art, while fitting a very tight hardware resource budget with 0.004\,mm${}^2$ and 0.0025\,mm${}^2$ of silicon area in UMC 65\,nm at 600\,MHz and 0.8\,word/cycle. The frequency can be pushed to 1.5\,GHz with a slight area increase of 25\%. 

We further show the proposed method works well not for various data formats and precisions, that the compression ratios are achieved reliably across many images with outliers only going down to 15\% below the average ratio at the 1st percentile. The same method is also applicable for the compression of gradient maps during training, achieving compression rates again more than 20\% higher than achieved for the feature maps in the forward pass. 
}

\bibliographystyle{IEEEtran}


\begin{IEEEbiography}[{\includegraphics[width=1in,height=1.25in,clip,keepaspectratio]{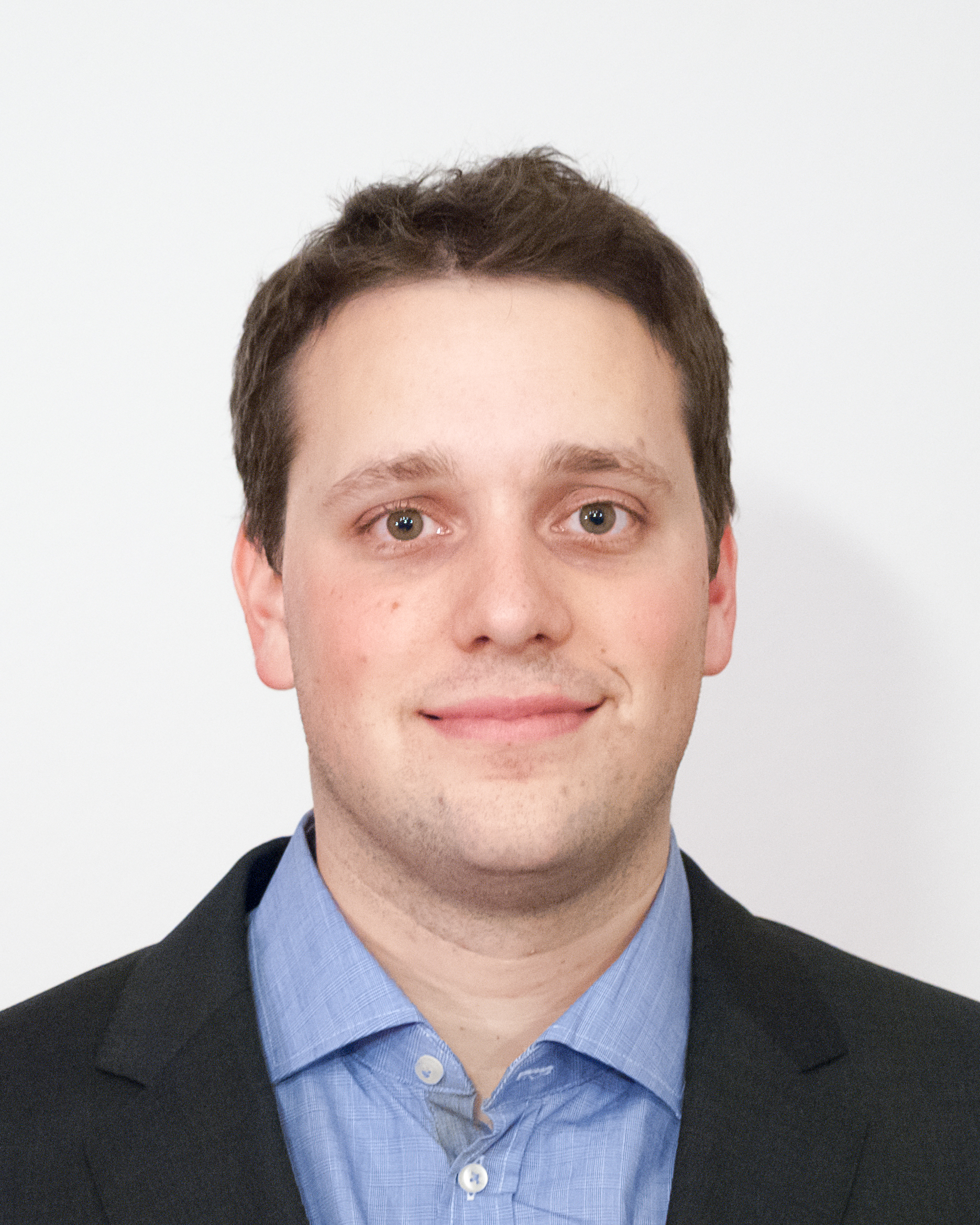}}]{Lukas Cavigelli} received the B.Sc., M.Sc., and Ph.D. degree in electrical engineering and information technology from ETH Zürich, Zürich, Switzerland in 2012, 2014 and 2019, respectively. He has since been a postdoctoral researcher with ETH Zürich. 
His research interests include deep learning, computer vision, embedded systems, and low-power integrated circuit design. He has received the best paper award at the VLSI-SoC and the ICDSC conferences in 2013 and 2017, the best student paper award at the Security+Defense conference in 2016, and the Donald O. Pederson best paper award (IEEE TCAD) in 2019. 
\end{IEEEbiography}
\begin{IEEEbiography}[{\includegraphics[width=1in,height=1.25in,clip,keepaspectratio]{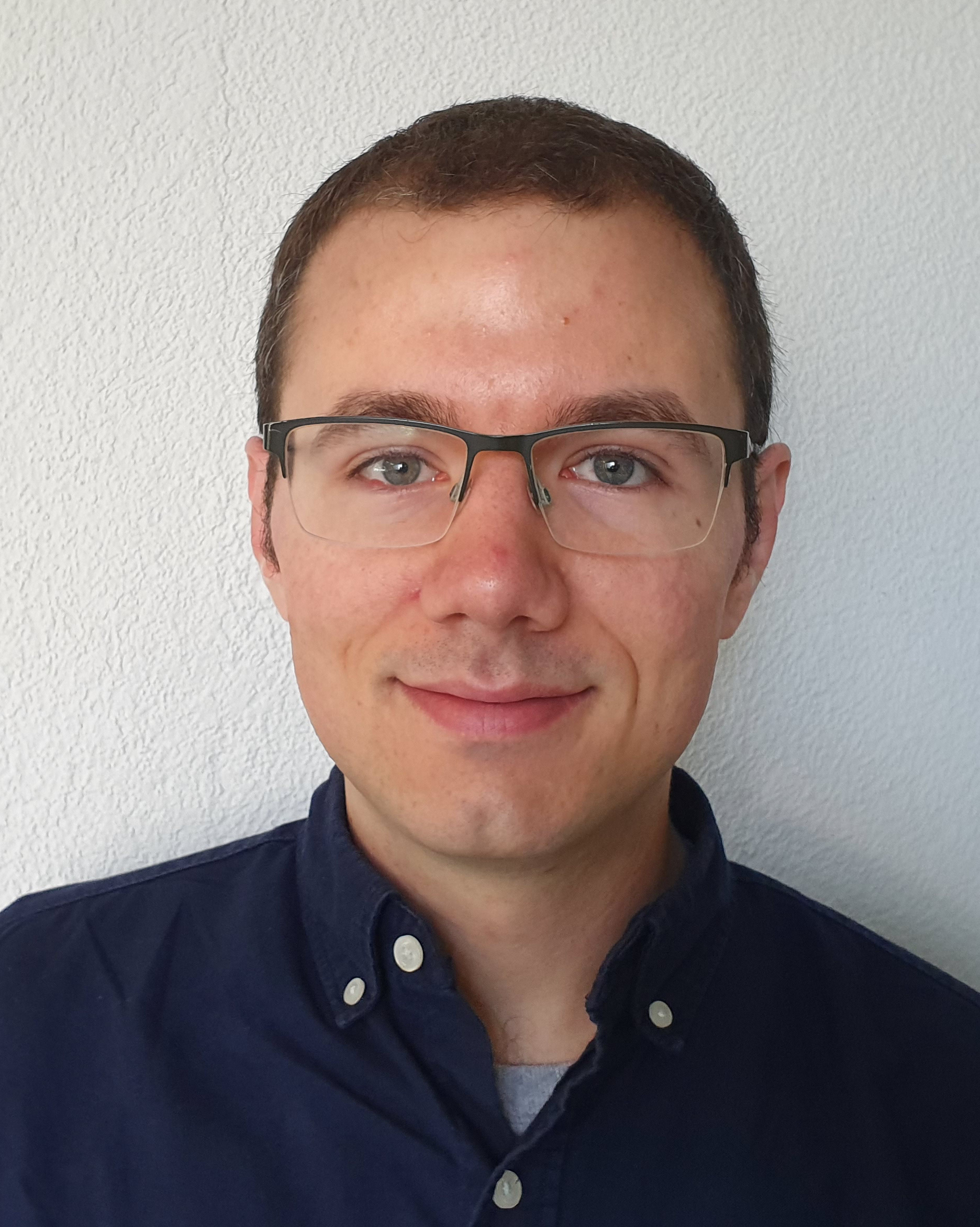}}]{Georg Rutishauser} received his B.Sc. and M.Sc. degrees in Electrical Engineering and Information Technology from ETH Zürich in 2015 and 2018, respectively. He is currently pursuing a Ph.D. degree at the Integrated Systems Laboratory at ETH Zürich. His research interests include algorithms and hardware for reduced-precision deep learning, and their application in computer vision and embedded systems.
\end{IEEEbiography}
\begin{IEEEbiography}[{\includegraphics[width=1in,height=1.2in,clip,keepaspectratio]{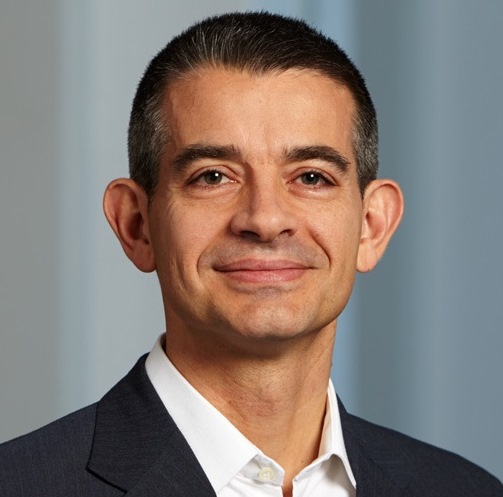}}]{Luca Benini} is the Chair of Digital Circuits and Systems at ETH Zürich and a Full Professor at the University of Bologna. He has served as Chief Architect for the Platform2012 in STMicroelectronics, Grenoble. Dr. Benini's research interests are in energy-efficient system and multi-core SoC design.  He is also active in the area of energy-efficient smart sensors and sensor networks. 
He has published more than 1'000 papers in peer-reviewed international journals and conferences, four books and several book chapters. He is a Fellow of the ACM and of the IEEE and a member of the Academia Europaea. 
\end{IEEEbiography}

\end{document}